\documentclass{article}

\usepackage[preprint]{neurips_2025}

\usepackage[utf8]{inputenc} 
\usepackage[T1]{fontenc}    
\usepackage{hyperref}       
\usepackage{url}            
\usepackage{booktabs}       
\usepackage{amsfonts}       
\usepackage{nicefrac}       
\usepackage{microtype}      
\usepackage{xcolor}         

\usepackage{amsmath,amsfonts,bm}

\def\eqref#1{equation~\ref{#1}}

\def\1{\bm{1}}

\DeclareMathAlphabet{\mathsfit}{\encodingdefault}{\sfdefault}{m}{sl}
\SetMathAlphabet{\mathsfit}{bold}{\encodingdefault}{\sfdefault}{bx}{n}

\usepackage{xcolor}
\usepackage{hyperref}
\usepackage{url}
\usepackage{multirow}
\usepackage{booktabs}
\usepackage{pifont}
\usepackage{xcolor}
\usepackage{graphicx}
\usepackage{amsmath}
\usepackage{algorithm}
\usepackage{mathtools}
\usepackage{algpseudocode}
\usepackage{listings}
\usepackage{wrapfig}
\usepackage{subcaption}
\usepackage{colortbl}
\usepackage{amssymb}
\usepackage{bbm}
\usepackage{gensymb}
\usepackage{pifont}
\usepackage{multicol}
\usepackage{xspace}
\usepackage{wrapfig}

\usepackage{enumitem}           

\newcommand{\paragraphVspace}

\newcommand{\name}{\texttt{VideoLifter}\xspace}

\title{VideoLifter: Lifting Videos to 3D with Fast and Efficient Hierarchical Stereo Alignment}

\author{
  \textbf{Wenyan Cong}$^{1}$,
  \textbf{Hanqing Zhu}$^{1}$,
  \textbf{Kevin Wang}$^{1}$,
  \textbf{Jiahui Lei}$^{2}$,
  \textbf{Colton Stearns}$^{3}$,\\
  \textbf{Yuanhao Cai}$^{4}$,
  \textbf{Leonidas Guibas}$^{3}$,
  \textbf{Zhangyang Wang}$^{1*}$,
  \textbf{Zhiwen Fan}$^{1*}$\\
  $^1$UT Austin\quad
  $^2$UPenn\quad 
  $^3$Stanford\quad
  $^4$JHU\quad\\
  \textbf{\textcolor{magenta}{Project Website}: \href{https://videolifter.github.io}{https://videolifter.github.io}}
}

\begin{document}
\maketitle
\begin{abstract}


Efficiently reconstructing 3D scenes from monocular video remains a core challenge in computer vision, vital for applications in virtual reality, robotics, and scene understanding. Recently, frame-by-frame progressive reconstruction without camera poses is commonly adopted, incurring high computational overhead and compounding errors when scaling to longer videos.
To overcome these issues, we introduce \name, a novel video-to-3D pipeline that leverages a local-to-global strategy on a fragment basis, achieving both extreme efficiency and SOTA quality.
Locally, \name leverages learnable 3D priors to register fragments, extracting essential information for subsequent 3D Gaussian initialization with enforced inter-fragment consistency and optimized efficiency. 
Globally, it employs a tree-based hierarchical merging method with key frame guidance for inter-fragment alignment, pairwise merging with Gaussian point pruning, and subsequent joint optimization to ensure global consistency while efficiently mitigating cumulative errors. 
This approach significantly accelerates the reconstruction process, reducing training time by over 82\% while holding better visual quality than SOTA methods.

\end{abstract} 
\begin{figure}[htb!]
  \centering
  \includegraphics[width=0.8\columnwidth]{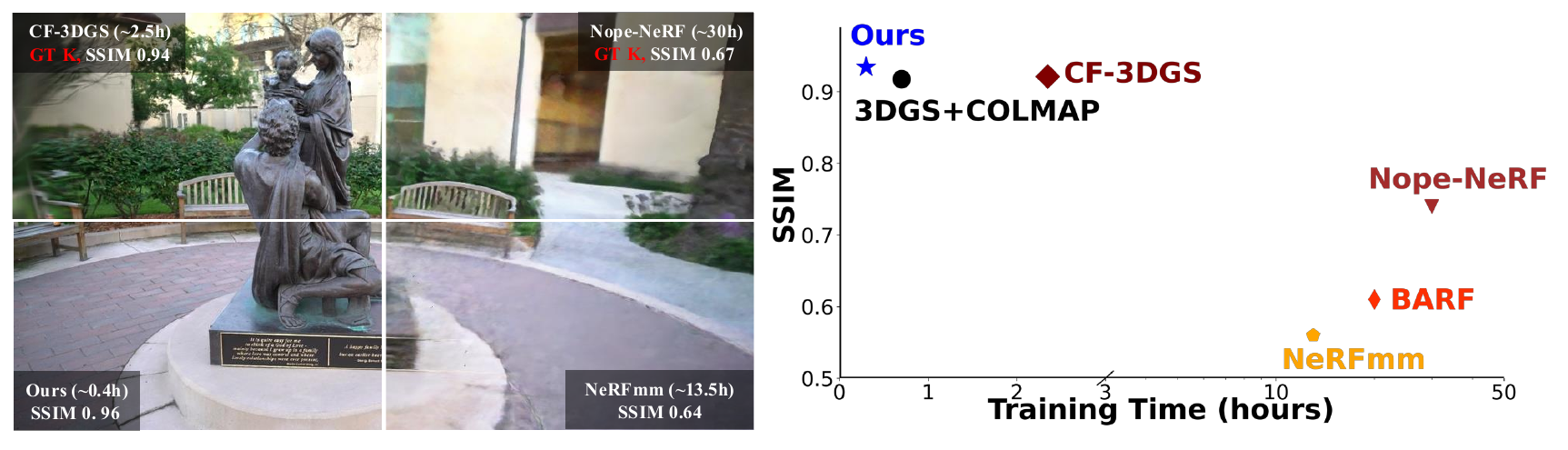}
    \captionof{figure}{
    \small
    \textbf{Novel View Synthesis and Training Time Comparisons}. \name does not require precomputed camera parameters (i.e., camera intrinsics K from COLMAP), reduces the training time required by the most relevant baseline CF-3DGS~\cite{fu2023colmap} by 82\% while improving image quality (SSIM).}
    \label{fig:teaser}
\end{figure}

\vspace{-7mm}
\section{Introduction}
\label{sec:intro}
Reconstructing 3D scenes from image observations is a longstanding problem in computer vision, with applications spanning AR/VR, video processing, and autonomous driving.
Recently, reconstructing 3D scenes from a single video (video-to-3D) has gained significant traction.
This trend is driven by two factors: the increasing accessibility of handheld capture devices, making video capture more practical for non-professional users, and recent advancements in high-fidelity 3D reconstruction methods such as Neural Radiance Fields (NeRF)~\cite{mildenhall2021nerf} and 3D Gaussian Splatting (3D-GS)~\cite{kerbl20233d}. 

Most video-to-3D reconstruction methods heavily depend on Structure-from-Motion (SfM)~\cite{schonberger2016structure} to generate initial sparse reconstructions, providing essential components like camera poses, intrinsics, and the initial point cloud to build dense 3D models using NeRF or 3DGS.
However, \textit{when applied to video data, SfM is often unreliable or even infeasible} (\textbf{Issue \ding{202}}), because it relies on photometric assumptions that frequently break down in low-texture or challenging lighting conditions~\cite{meuleman2023progressively, moulon2013global, schonberger2016structure},
though some works improve SfM for some specific conditions~\cite{lindenberger2021pixel}.
In response, recent methods~\cite{fu2023colmap, bian2023nope, lin2021barf, wang2021nerf} have shifted toward jointly optimizing camera poses and scene representations rather than relying solely on SfM-based initializations. But these approaches still depend on accurate camera intrinsics from SfM, limiting their applicability in in-the-wild video scenarios.

More importantly, those SfM-free video-to-3D methods typically reconstruct scenes incrementally from a canonical view, with two critical issues.
First, they are \textit{slow and inefficient} (\textbf{Issue \ding{203}}) due to an iterative, frame-by-frame approach that re-optimizes the entire sequence with each new frame, thereby prolonging training times ($>$ 2 hours) and complicating the handling of complex trajectories,
especially given the video setup, not a few images. 
Naively using non-incremental InstantSplat~\cite{fan2024instantsplat} cannot handle video data with an Out-of-memory (OOM) issue, which cannot scale to many frames (See Tab.~\ref{tab:oom}).
Second, they are \textit{susceptible to incremental errors} (\textbf{Issue \ding{204}}), as the frame-by-frame approach tends to accumulate errors over long video sequences.

To address these issues, we propose \name, a novel video-to-3D reconstruction pipeline that achieves a \textbf{5$\times$ speed-up} and \textbf{enhanced novel view-synthesis quality} compared to state-of-the-art methods, as demonstrated in Fig.~\ref{fig:teaser}.
We effectively adopt the \textbf{local-to-global} idea stream to handle long-sequence videos on a fragment basis and then subsequently merge fragments into a final, globally consistent 3D scene. 
Our pipeline is driven by two key innovations that make the local-to-global concept workable with significantly boosted efficiency (Issue \ding{203}) and much-reduced incremental errors (Issue \ding{204}) on video-to-3D.
First, in the \textbf{Fragment Registration with Learned 3D Priors} (Local) stage, we extract essential information (e.g., pointmaps and local camera poses for 3D Gaussian initialization) from each fragment by leveraging pretrained prior models such as MASt3R~\cite{leroy2024grounding} to address Issue \ding{202}.  
Rather than naively using 3D priors to initialize 3D Gaussians, like InstantSplat (MASt3R + 3D-GS)~\cite{fan2024instantsplat}, 
we improve \textit{efficiency} by (1) enforcing inter-fragment consistency via solely considering the key frames, solved on an efficient subgraph instead complete graph, and (2) extracting only the essential parameters (6-dimensional quaternion pose and 1-dimensional scale) for each view within fragment, thereby avoiding costly global optimization of full point maps.
Second, in the \textbf{Hierarchical Gaussian Alignment} (Global) stage, we merge fragments through a tree-based hierarchical framework that employs key frame guidance for inter-fragment alignment, pairwise merging with Gaussian point pruning, and subsequent joint optimization to ensure global consistency and mitigate cumulative errors efficiently.
Overall, our main contributions are as follows:
\begin{itemize}[leftmargin=1em]
\item We introduce \name, an efficient, high-quality, and robust video-to-3D reconstruction framework with a local-to-global strategy.
\item Our \textbf{fragment registration with learned 3D priors} efficiently extracts essential representations for subsequent dense 3D-GS with several key \textit{efficiency-driven optimizations} along with \textit{learned 3D priors} to remove reliance on traditional module SfM. 
\item Our \textbf{hierarchical 3D Gaussian alignment} minimizes incremental errors through three well-designed iterative stages, ensuring both accuracy and efficiency.
\item Extensive experiments on the Tanks and Temples and CO3D-V2 datasets demonstrate that \name significantly enhances training efficiency, with more than 5× speed improvements, and improves rendering quality compared to state-of-the-art methods. 

\end{itemize}

\section{Related Works}
\vspace{-10pt}
\paragraph{\textbf{3D Representations for Novel View Synthesis}}
3D reconstruction for high-quality novel view synthesis generates unseen views of a scene or object from a set of images~\cite{avidan1997novel,mildenhall2019local}.
After the seminal NeRF work~\cite{mildenhall2021nerf}, a wave of unstructured radiance field methods has emerged~\cite{kerbl20233d, xu2022point}, each adopting different scene-representation primitives. 
Among these, 3D-GS~\cite{kerbl20233d} stands out with impressive performance in efficiently reconstructing complex, real-world scenes with high fidelity. 
Both NeRFs and 3DGS rely on carefully captured sequential video or multi-view images to ensure sufficient scene coverage, utilizing preprocessing tools like SfM software, e.g., COLMAP~\cite{schonberger2016structure}, to compute camera parameters and provide a sparse SfM point cloud as additional input. 

\vspace{-10pt}
\paragraph{Traditional Structure-from-Motion (SfM) }
Estimating 3D structure and camera motion is a well-explored challenge~\cite{furukawa2010towards,newcombe2011dtam,wu2011visualsfm}.
SfM has seen significant advancements across various dimensions.
Methods like \cite{lowe2004distinctive,MatchNet} focus on enhancing feature detection, \cite{snavely2011scene} introduces innovative optimization methods, \cite{gherardi2010improving,schonberger2016structure} explore improved data representations and more robust structural solutions.
Despite these advances, traditional SfM techniques remain vulnerable to issues such as low-texture regions, occlusions, moving objects, and lighting variations, limiting their overall robustness and performance.

\vspace{-10pt}
\paragraph{\textbf{Radiance Field without SfM}}
Inaccuracies from SfM can propagate through subsequent radiance field reconstruction, reducing overall quality. 
Various approaches have been proposed to eliminate the reliance on SfM by jointly optimizing camera parameters and scene representation, such as NeRFmm~\cite{wang2021nerf} and BARF~\cite{lin2021barf}.
GARF~\cite{jeong2021self} further simplifies the joint optimization and improve both efficiency and accuracy by using Gaussian-MLP models. SPARF~\cite{truong2023sparf} and TrackNeRF~\cite{mai2024tracknerf} introduces a method to simulate pose noise by injecting Gaussian noise into the camera parameters.
Recently, depth priors from monocular depth estimator are used to guide radiance fields optimization~\cite{bian2023nope,cheng2023lu,meuleman2023progressively,fu2023colmap}. More recent work, such as~\cite{jiang2024construct} and InstantSplat~\cite{fan2024instantsplat}, either supplement depth priors with other priors (e.g., image matching network), or integrate end-to-end stereo models like DUST3R/MASt3R to reduce the dependency on camera pose information. 
While these methods show promise in removing the SfM reliance, scaling them to a large number of views remains a challenge, such as the OOM issue (e.g., InstantSplat), drifting error (e.g., CF-3DGS), and unsatisfactory quality (e.g., InstantSplat).

\vspace{-10pt}
\paragraph{Comparison with Simultaneous Localization and Mapping (SLAM)}  
Although both SLAM and video-to-3D reconstruction process multiple views, their input conditions and end goals differ fundamentally.  
First, SLAM operates online, processing frames sequentially as they arrive, whereas video-to-3D has access to the entire sequence upfront. This offline setting enables pipelines to consider all frames together rather than a purely sequential approach.  
Second, our primary objective is novel view synthesis, generating photorealistic views from unseen viewpoints, which SLAM methods are not designed to support, but only to rerender the training set~\cite{sandstrom2024splat}.  
Hence, SLAM-based techniques are not directly applicable to the video-to-3D reconstruction problem.






\section{\name: An Efficient and Effective Video-to-3D Framework}

\subsection{Video-to-3D: Challenges and Our Design}

We first define the video-to-3D reconstruction problem and outline current issues in delivering an \textit{efficient and high-quality} reconstruction pipeline. We then present our high-level design philosophy that tackles these challenges.

\vspace{-10pt}
\paragraph{Video-to-3D Reconstruction} 
Given a sequence of $N$ unposed and uncalibrated images from a monocular video, denoted as $\mathcal{I} = \{I_i \in \mathbb{R}^{H \times W \times 3} \}_{i=1}^{N}$, 
\name aims to reconstruct the scene
\begin{wrapfigure}{r}{0.38\textwidth}
  \vspace{-1.5em}
  \centering
  \includegraphics[width=0.9\linewidth]{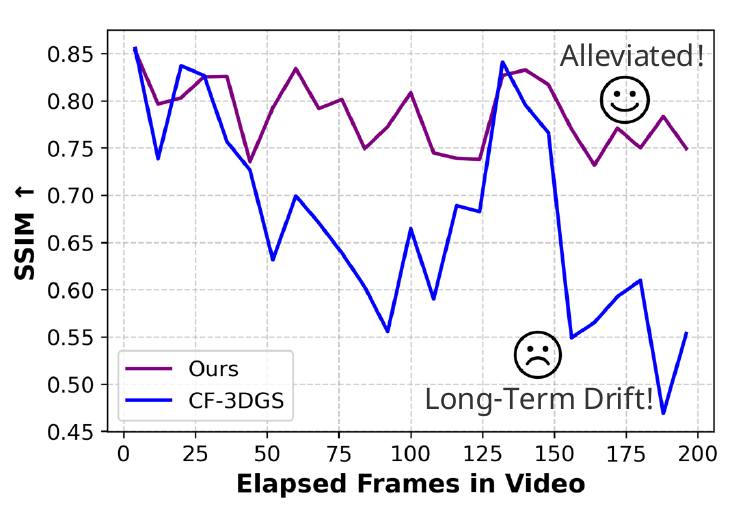}
  \vspace{-0.5em}
  \caption{\small CF-3DGS's frame-by-frame pipeline accumulates errors with long-term drift, while our method compresses drift error along \# frames. Results are tested on 247\_26441\_50907 from CO3D-V2.}
  \label{figure:drift}
  \vspace{-1.5em}
\end{wrapfigure}
using 3D Gaussians
$\mathcal{G}$ along with estimated camera intrinsics $K$ and extrinsics $\mathcal{T} = \{ T_i \in \mathbb{R}^{3 \times 4} \}_{i=1}^{N}$. 
We assume that all frames share a common intrinsic matrix, as they are from a single monocular video.

\vspace{-10pt}
\paragraph{Key Challenges}
While NeRF and 3D-GS have advanced 3D reconstruction, 
their variants remain suboptimal for video-to-3D reconstruction in terms of \textbf{speed} and \textbf{quality}.
Existing methods often adopt a frame-by-frame progressive reconstruction method, 
making them inherently \textit{slow} and \textit{prone to cumulative errors} (See long-term performance drift in Fig. \ref{figure:drift}) when processing videos. 
Furthermore, they typically rely on SfM to estimate camera intrinsic, which is \textit{unreliable} or \textit{even infeasible} for in-the-wild video sequences.

\vspace{-8pt}
\paragraph{Our High-level Framework}
To meet the critical need for efficiency and quality in long-sequence video to 3D reconstruction, we depart from conventional frame-by-frame or holistic optimization methods by embracing a hierarchical local-to-global design philosophy. 
Specifically, we process long video sequences on a \textit{fragment} basis and subsequently merge these fragments into a single, consistent 3D scene. 
Although the local-to-global concept is not new, adapting it to a video-to-3D reconstruction pipeline with 3DGS is new, with two key, unanswered challenges:


\colorbox{yellow!20}{%
        \parbox{\dimexpr\linewidth-2\fboxsep}{%
            \textit{How can we extract 3D reconstruction information \textbf{efficiently} and \textbf{reliably} from monocular video?}
        }%
    }

\colorbox{blue!10}{%
        \parbox{\dimexpr\linewidth-2\fboxsep}{%
            \textit{How can we merge fragments into a \textbf{high-quality} and \textbf{consistent} 3D scene without alignment issues?}%
        }
    }

\begin{figure*}[t!]
  \centering
  \includegraphics[width=0.95\linewidth]{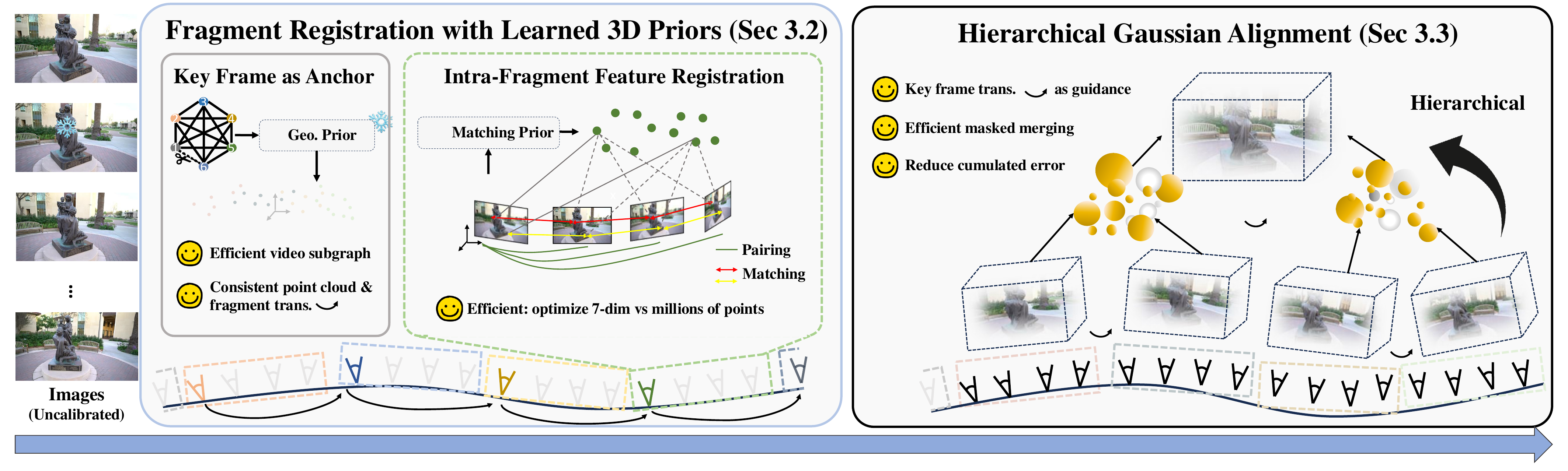}
\vspace{-5pt}
\caption{
\small \textbf{Network Architecture.} Given uncalibrated images, \name first employs learned priors for efficient fragment registration. The independently optimized 3D Gaussians from fragment are then hierarchically aligned into a globally coherent 3D representation.}
\label{fig:arc}
\vspace{-15pt}
\end{figure*}

In response, we propose \name, an \textbf{efficient} and \textbf{high-quality} video-to-3D reconstruction pipeline, as illustrated in Fig.~\ref{fig:arc}. 
Our method achieves SOTA performance in both speed and quality (see Fig.~\ref{fig:teaser}) through two-level innovations: \colorbox{yellow!20}{Local:} \textbf{Fragment Registration with Learned 3D
Priors} with efficiency-driven designs (Sec.~\ref{subsec:fragment}) and 
\colorbox{blue!10}{Global:} \textbf{Hierarchical Gaussian Alignment} to compress alignment error (Sec.~\ref{subsec:merge}).

\subsection{Fragment Registration with Learned 3D Priors}
\label{subsec:fragment}

First, we describe our method to \textit{efficiently} and \textit{reliably} extract essential 3D reconstruction information on a fragment basis and the way we enforce cross-fragment consistency.

\noindent\textbf{Process Description:}
We partition the input video sequence into $m$ disjoint windows of length $k$, which we refer to as \textbf{fragments}. 
For example, the $i$'th fragment is given by $\mathcal{I}^f_i = [I_{(i-1)k+1}, I_{(i-1)k+2}, \dots, I_{ik}]$, where $i \in [1,m]$.
Our objective is twofold: first, to extract essential information for subsequent intra-fragment 3D reconstruction (e.g., point cloud for Gaussian initialization, along with coarse local camera extrinsics and intrinsics); and second, to obtain inter-fragment information necessary for future local-to-global merging. 

\noindent\underline{\textit{Fragmentation Method Choice:}}
In this work, we use a straightforward fragmentation strategy by uniformly dividing the frames into disjoint windows. 
Although simple, this approach has proven effective on benchmarks. 
More sophisticated methods, such as using frame-to-frame similarity to guide fragmentation, could further enhance \name, e.g., in videos with abrupt view changes during capture, but are orthogonal to our core contributions.


\noindent\textbf{Fragment-level Challenges:}
Naively applying existing methods for the fragment level (e.g., those in LocalRF~\cite{meuleman2023progressively} or CF-3DGS~\cite{fu2023colmap}) is neither efficient enough nor does it adequately prepare for future merging.
Challenge~\ding{192}: SfM is heavily relied for NeRF/3D-GS, while it is not always available or reliable—especially in our video-to-3D reconstruction with varying conditions.
Challenge~\ding{193}: A critical issue in the local-to-global paradigm is ensuring that fragments can be merged without incurring significant alignment errors. 
Challenge~\ding{194}: Beyond the inherent efficiency benefits of a local-to-global design, further enhancements in efficiency are necessary at the video setup.

\noindent\textbf{Learned 3D priors} (Challenge \ding{192}): 
3D-GS needs \textit{point cloud} for initialization and \textit{camera pose} for optimization.
However, in long-sequence video settings, traditional SfM methods (e.g., COLMAP) are often unavailable or unreliable.
While CF-3DGS employs monocular depth estimation (a geometric prior) on each view to obtain a point cloud, it introduces scale issues that necessitate additional optimization during 3D-GS training.

Inspired by recent work on replacing SIFT with NN-based method, e.g., LoFTR~\cite{sun2021loftr}/GIM \cite{shen2024gim}, 
\begin{wraptable}{r}{0.55\textwidth}
\vspace{-10pt}
  \centering
  \small
  \resizebox{0.98\linewidth}{!}{
\begin{tabular}{c|c|cccc}
\hline
Matching & Geometric & SSIM & PSNR & LPIPS & ATE\\ \hline
LoFTR~\cite{sun2021loftr} & Metric3Dv2 \cite{hu2024metric3d} &  0.9238 & 31.30 & 0.0757 & 0.005\\
MASt3R & MASt3R & 0.9347 & 31.59 & 0.0730 & 0.004 \\\hline
\end{tabular}}
\caption{\small Comparison of prior models on Tanks and Temples.}
\label{tab:alter_prior}
\vspace{-10pt}
\end{wraptable}
we abandon reliance on SfM and instead leverage learned 3D priors from large-scale pretrained foundation models.
In this work, we use MASt3R~\cite{leroy2024grounding} as prior model since it seamlessly integrates both geometric and matching cues. We emphasize that our video-to-3D pipeline are not exclusively tailored to MASt3R; rather, our \name is flexible and can incorporate any model that provides robust geometric and matching priors. For example, in Tab.~\ref{tab:alter_prior}, we demonstrate that \name performs well with both MASt3R and alternative approaches (e.g. LoFTR~\cite{sun2021loftr} for matching cues, Metric3Dv2 \cite{hu2024metric3d} for geometric cues). We choose MASt3R for its simplicity and efficiency.

\noindent\textbf{Key Frames as Anchor} (Challenge \ding{193}, \ding{194}): 
In a fragment-based approach, ensuring inter-fragment consistency is critical for high-quality 3D reconstruction. 
To address this, 
we propose to ``anchor" each fragment with a \textbf{key frame}, set to its first frame, $I_{(i-1)k+1}$ and \textit{enforce consistency only among these key frames}.
This strategy simplifies the consistency problem by limiting the problem scale in $\frac{N}{k}$ key frames instead of considering whole $N$ frames, thereby \textit{improve efficiency} and \textit{reduce complexity}. 

Key frame consistency is enforced at the \textit{point-cloud} level
by generating a globally optimized dense point map, along with transformation matrices across adjacent fragments $\{T^f_{i\rightarrow i+1}\}_{i=1}^{m-1}$.
To achieve this, we follow a procedure similar to that in MASt3R~\cite{leroy2024grounding}, optimizing: 
\begin{equation}
    \small
(\tilde{\boldsymbol{P}}^{*}, \boldsymbol{T}_e^{*}) = \arg \min_{\tilde{\boldsymbol{P}}, \boldsymbol{T}, \sigma} \sum_{e \in \mathcal{E}} \sum_{v \in e} \sum_{i=1}^{HW}
    \boldsymbol{O}_{v,e}^i \left\Vert \tilde{\boldsymbol{P}}_{v}^{i} - \sigma_e \boldsymbol{T}_e \boldsymbol{P}_{v,e}^{i} \right\Vert.
\end{equation}
where for each image pair \( e = (v, u) \in \mathcal{E} \), $\sigma_e$ is scale factor, \(\boldsymbol{P}_{v,e}\) and \(\boldsymbol{O}_{v,e}\) is the pointmap and confidence map of $v$, respectively.
\textit{However, MASt3R builds a complete graph for this optimization, resulting in a complexity of  $\mathcal{O}((\frac{N}{k})^2)$, which becomes inefficient enough for long video sequences.}

\underline{\textit{To enhance efficiency}} rather than naively using MASt3R-like method, we propose to build a more efficient sub-graph that only builds edges between the key frames and their four closest neighboring frames.
\textit{This design is motivated by the observation that neighboring segments share greater co-visibility; hence, edges between key frames with distant neighbors can be safely pruned.}
This sub-graph greatly reduces the optimization complexity to $\mathcal{O}(4\frac{N}{k})$, which scales linearly with \#frames $N$, while still showing high end-to-end 3D reconstruction quality.

\noindent\textbf{Efficient intra-Fragment Feature Registration} (Challenge \ding{194}): 
Finally, we aim to obtain an initial estimate of the local camera poses and pointmaps along with depth scale factors within each non-overlapping fragment, which can accelerate and boost the quality of the subsequent 3D Gaussian construction.
\textit{A naive solution is to follow InstantSplat~\cite{fan2024instantsplat} to use MASt3R to solve the intra-fragment problem, but it requires optimizing millions of points and camera poses, making it inefficient.}

\underline{\textit{To enhance efficiency}}, we use the pre-obtained key frame information and obtain the needed information only considering pairwise relationships between key frame and all subsequent frames in the same fragment.
In this way, we only need to solve for 6-dimensional camera poses (in quaternion format) and a 1-dimensional scale factor for each view.
We found it sufficient to maintain end-to-end reconstruction quality with high efficiency. Moreover, this simplified, non-sequential matching approach can reduce the incremental errors that are commonly encountered in sequential matching(See Fig.~\ref{figure:drift}).

Take first fragment \( \mathcal{I}^f_1 = \{I_{1}, \dots, I_{k}\} \) as an example.

\noindent\textit{Camera pose}: 
We refine the relative camera poses within the fragment using initial pairwise estimates. 
First, we identify the intersection of 2D correspondences between the key frame  \( I_{1} \) and each subsequent frame from index 2 to \( k \). 
This process yields a consistent set of correspondences across all frames in the fragment. Using these intersected 2D correspondences, we retrieve the corresponding 3D positions from the key frame, which were previously optimized during key frame processing.
These 3D-2D correspondences are then input to PnP-RANSAC~\cite{fischler1981random}, refining the camera poses to ensure alignment with consistent 3D points across views within the fragment.
Only a 6-dim quaternion pose is optimized instead of directly optimizing pointmaps.

\noindent\textit{Scale factor}:
Scale variations may persist within the point clouds of the fragment due to independent inference. 
To address this, intersected 3D points from the key frame are utilized for scale estimation across all image pairs. 
Specifically, for each image pair between \( I_{1} \) and \( \{I_i\}_{i=2}^{k} \), the corresponding 3D point positions (with a total of \( P \) points) are retrieved. A one-degree-of-freedom (1-DoF) scale factor is computed between the intersected 3D points in the current pair and those from the key frame:
\begin{equation}
\label{eq:scaling_median} 
\small 
s_i = \operatorname{median} ( \left\{ \|\mathbf{p}_n^{(1)}\| / \|\mathbf{p}_n^{(i)}\| \right\}_{n=1}^{P} ), 
\end{equation}
which can be solved analytically (i.e., by taking the median) with \textit{no optimization need}.
This scale factor is applied to the dense pointmaps of \(I_i\) in the subsequent stage, ensuring that the point clouds within the fragment are locally aligned and maintain a consistent scale relative to the key frame.

By solving a simplified optimization problem rather than naively employing MASt3R global optimization as in InstantSplat, our method achieves a processing time of 2.97 seconds compared to 10.33 seconds—a 3× reduction on computational time within fragment $k$=4.

\noindent\textbf{Overall Efficiency/Quality Improvement with Our Novel Fragment Registration:}
In contrast
\begin{wraptable}{r}{0.55\textwidth}
\vspace{-10pt}
  \centering
  \small
  \resizebox{0.98\linewidth}{!}{%
    \begin{tabular}{c|cc|cc}
      \toprule
      \multirow{2}{*}{\# Views} 
        & \multicolumn{2}{c|}{MASt3R (Global Optimization)} 
        & \multicolumn{2}{c}{VideoLifter (Sec.~\ref{subsec:fragment})} \\
      & \begin{tabular}{c}Time \\ (min) \end{tabular} 
        & \begin{tabular}{c}Peak GPU\\Mem (GB)\end{tabular} 
        & \begin{tabular}{c}Time \\ (min) \end{tabular}
        & \begin{tabular}{c}Peak GPU\\Mem (GB)\end{tabular} \\
      \midrule
      32  & 11  & 4.75 & 1.7 & 4.45 \\
      48  & 33  & 8.86 & 2.6 & 5.43 \\
      64  & 63  & 14.44 & 3.5 & 6.38 \\
      128 & OOM & OOM   & 7.9 & 16.9 \\
      \bottomrule
    \end{tabular}%
  }
  \caption{\small Time and peak GPU memory usage on an A6000 for varying view counts: Direct using MASt3R global optimization vs.\ our Sec.~\ref{subsec:fragment}.}
  \label{tab:oom}
  \vspace{-10pt}
\end{wraptable}
to naively applying MASt3R's global optimization as InstantSplat~\cite{fan2024instantsplat} to compute point clouds and camera poses for all images, our method significantly enhances \textbf{efficiency}. As in Tab.~\ref{tab:oom}, their global optimization approach runs out of memory beyond 64 views and incurs up to an 18$\times$ longer inference time. 
Moreover, Tab.~\ref{tab:ablation} demonstrates that replacing our fragment registration with  MASt3R initialization yields lower-quality results. 
This is primarily because it trys to solve a more complex global problem with less accurate pointmaps/poses outputs, whereas our approach eases optimization complexity by providing a more robust initialization for subsequent 3D-GS, delivering \textbf{better quality}.

\subsection{Hierarchical Gaussian Alignment}
\label{subsec:merge}

In this stage, we perform dense 3D scene reconstruction using Gaussian Splatting. First, we construct \textit{local 3D Gaussians} within each fragment, and then merge these local models via \textit{hierarchical Gaussian alignment}. 
The key design question is \textit{how to construct a globally coherent 3D scene while preserving local scene details without incurring significant alignment errors}.



\vspace{-8pt}
\paragraph{Local 3D Gaussian Construction:}
We initialize a set of Gaussians, denoted as $\mathcal{G}^f = \{ G^f_i \}_{i=1}^{m}$, where $G^f_i$ is independent initialized and optimize from fragment $\mathcal{I}^f_i$.

\noindent\textit{Guassian initialization}: 
In the local fragment registration step, we obtain the key frame's dense point cloud, along with the relative poses and scale factors for the other frames, which can be used to obtain entire point map within fragment.
To initialize $G^f_i$,
we then assign a Gaussian to each point in the pixel-wise point cloud, setting its attributes as: color based on the corresponding pixel, center at the 3D point location, opacity adhering to the 3D-GS protocol~\cite{kerbl20233d}, and scaling such that it projects as a one-pixel radius in the 2D image (by dividing the depth by the focal length). We set Gaussians as isotropic to reduce the degrees of freedom in Gaussian training.

\noindent\textit{Further refinement}:
The initial camera poses and point cloud positions may contain minor 
\begin{wrapfigure}{r}{0.5\textwidth}
  \vspace{-1.2em}
  \centering
  \includegraphics[width=1\linewidth]{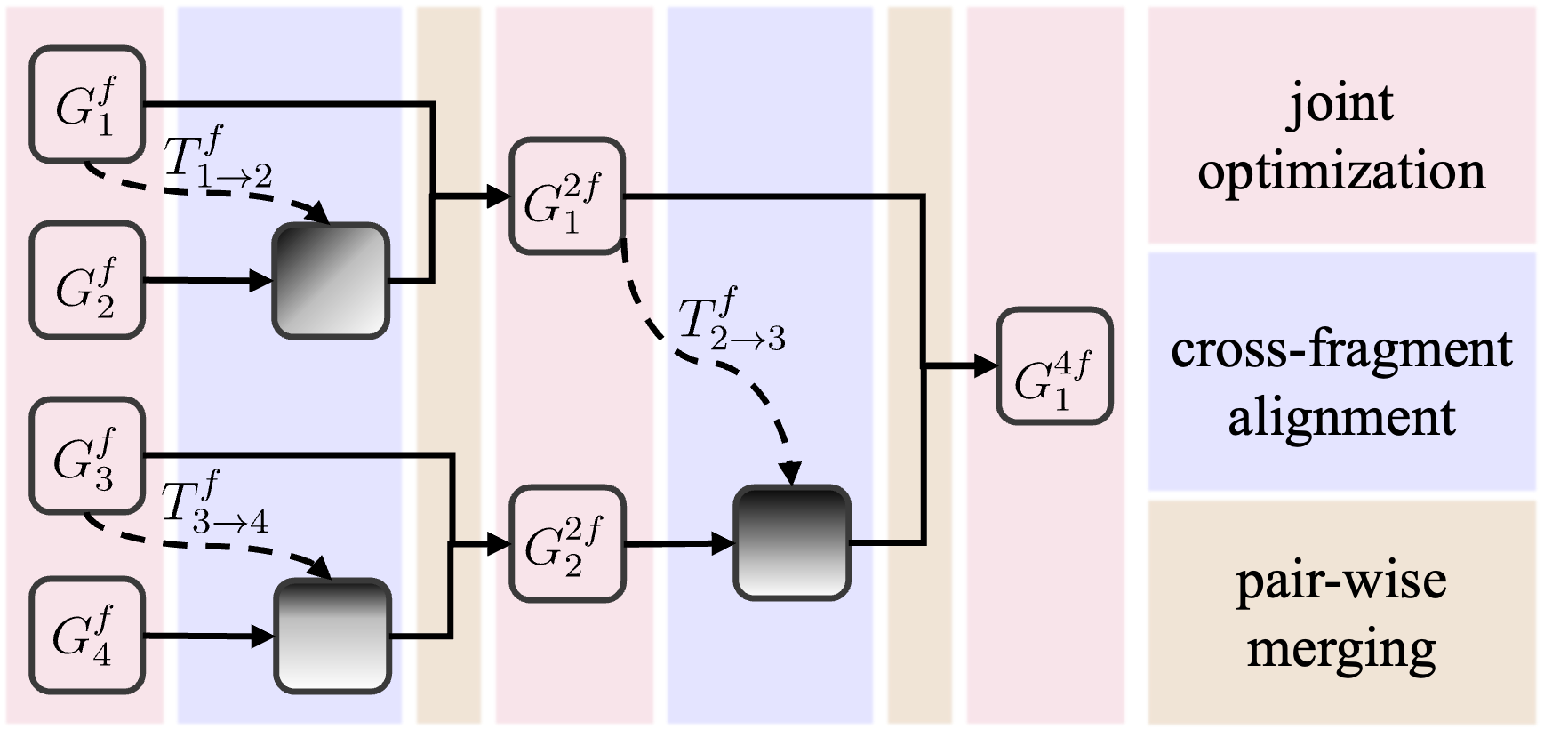}
\caption{\small \textbf{Hierarchical Gaussian Alignment.} The process iteratively performs three stages: 1) joint optimization of camera poses and local Gaussians (pink), 2) cross-fragment alignment for new local Gaussian (purple), and 3) visibility masking and pairwise merging of local Gaussians (yellow), until a globally consistent scene reconstruction is achieved.}
\label{fig:merge}
\vspace{-20pt}
\end{wrapfigure}
inaccuracies, we further refine them through joint optimization of camera poses and Gaussian parameters. Specifically, for each local Gaussian in \( G^f_i \), we randomly sample frames within the fragment, render the current Gaussians into sampled frame, and backpropagate gradient updates to the Gaussian positions, colors, scales, opacities, and camera poses.


\vspace{-8pt}
\paragraph{Hierarchical Gaussian Alignment:}
Next, we merge the local fragment-level 3D Gaussian sets $\mathcal{G}^f = \{ G^f_i \}_{i=1}^{m}$ to the final consistent 3D scene.
Naive pairwise progressive merging poses Challenge~\ding{192}: an excessive number of Gaussians for optimization and \ding{193}: inconsistencies among various local Gaussian sets.
To avoid these issues, we propose a tree-based hierarchical pipeline ( Fig.~\ref{fig:merge}) that iteratively performs three key processes.


\noindent1) \textcolor{blue!50}{Inter-Fragment Alignment with Key Frame Guidance} (Fig.~\ref{fig:merge} purple):
To merge two independently optimized fragments (e.g., \( G^f_1 \) and \( G^f_2 \)), we first perform \textbf{cross-fragment alignment} to ensure a faithful merging by using \( G^f_1 \) as the reference coordinate system. 
In each fragment, the key frame, i.e., the first frame, is assigned an identity pose, and the remaining frames are defined by their relative poses to this key frame.
In Section~\ref{subsec:fragment}, we enforce consistency between key frames and obtain the initial transformation $T^f_{1\rightarrow 2}$. We then use this information as a guide to compute the camera poses for the novel frames covered by \( G^f_2 \). By enforcing photometric loss on the next novel view while freezing all parameters in \( G^f_1 \), we could further optimize $T^f_{1\rightarrow 2}$ into $T^{f*}_{1\rightarrow 2}$. Then, we align the Gaussians in \( G^f_2 \) with the coordinate system of \( G^f_1 \) by using ${T^{f*}_{1\rightarrow 2}}^{-1}$. As shown in Tab.~\ref{tab:ablation}, omitting key frame guidance leads to prolonged optimization and degraded performance.


\noindent2) \textcolor{brown}{Pair-wise merging with visibility-mask-driven Gaussian pruning} (Fig.~\ref{fig:merge} yellow):
To avoid duplicating Gaussians in regions where \( G^f_1 \) already provides adequate scene reconstruction, with $\mathbf{p}$ (the pixel position on image plane), we use a \textbf{visibility mask} to determine areas that \( G^f_1 \) could faithfully reconstruct:
$
M(\mathbf{p}) = \text{Conf}(\mathbf{p}) > \beta + D(\mathbf{p}) > 0.
$
where $D(\mathbf{p})$ is rendered depth.
 $\text{Conf}(\mathbf{p})=\sum_i\alpha_i(\mathbf{p}) \prod_{j=1}^{i-1} \left(1 - \alpha_j(\mathbf{p})\right)$ is the rendered confidence, e.g. if a pixel contains information from
the current gaussian, and threshold $\beta$ determines the masking criteria.

\begin{wrapfigure}{r}{0.48\textwidth}
  \vspace{-1.5em}
  \centering
\includegraphics[width=0.95\linewidth]{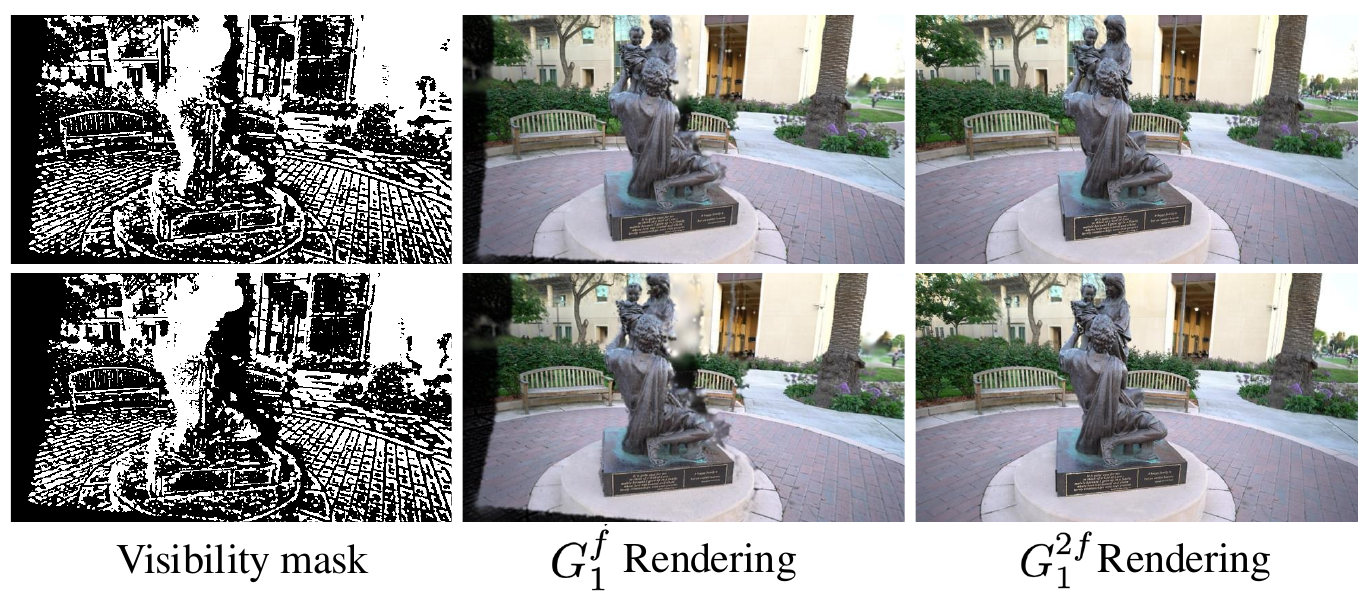}
\vspace{-0.5em}
\caption{\small \textbf{Visibility Mask} showing rendering \( G^f_1 \) into next two novel frames in \( G^f_2 \). White region denotes faithful reconstruction using \( G^f_1 \), while black represents pixels unseen from \( G^f_1 \). With visibility mask, we select complementary gaussians from \( G^f_2 \), merging them with \( G^f_1 \) into \( G^{2f}_1 \).}
\label{fig:mask}
\vspace{-20pt}
\end{wrapfigure}

Fig.~\ref{fig:mask} shows that the visibility mask effectively identifies regions where Gaussians from \( G^f_1 \) could provide sufficient depth and confidence. Importantly, the mask remains robust to occlusions, even when covered by large Gaussians.
As our initialization is pixel-wise point clouds, the inverse of the visibility mask can be directly applied to \( G^f_2 \) to select Gaussians that complement the missing regions of \( G^f_1 \). These selected Gaussians inherit previously optimized parameters, ensuring seamless integration into a unified representation.

\noindent3) \textcolor{purple!50}{Joint optimization} (Fig.~\ref{fig:merge} pink):
After merging pair of local Gaussians, further \textbf{joint optimization} is needed to ensure the merged Gaussian set \( G^{2f}_1 \) meets our consistency objectives. We jointly optimize Gaussian properties and camera poses for \( G^{2f}_1 \). Specifically, we randomly sample frames within \( [I_1, I_8] \), render all Gaussians into each frame, and backpropagate gradients to Gaussian positions, colors, scales, opacities, and camera poses.

\section{Experiments}

\subsection{Experimental Setup}

\noindent\textbf{Datasets:}
We carry out comprehensive experiments on various real-world datasets, including Tanks and Temples~\cite{knapitsch2017tanks}, CO3D-V2~\cite{reizenstein2021common}. 
For Tanks and Temples, following NoPe-NeRF~\cite{bian2023nope} and CF-3DGS~\cite{fu2023colmap}, we assess both novel view synthesis quality and pose estimation accuracy across 8 scenes, spanning both indoor and outdoor environments. 
In each case, we use 7 out of every 8 frames from the video clips as training data and evaluate the novel view synthesis on the remaining frame except Family. 
For CO3D-V2, containing thousands of object-centric videos where the camera orbits the object, recovering camera poses is significantly more challenging due to the complex and large camera motions. \textit{We follow the experimental settings of CF-3DGS~\cite{fu2023colmap}} to select the same 10 scenes from different object categories and apply the same procedure to divide training and testing sets.

\vspace{-2pt}
\begin{table*}
\centering
\resizebox{0.95\linewidth}{!}{
\begin{tabular}{lccccccc}
\toprule
& Camera Param. & Train Time & SSIM $\uparrow$ & PSNR $\uparrow$ & LPIPS $\downarrow$ & ATE $\downarrow$ \\
\hline
COLMAP+3DGS  & GT K \& Pose  & $\sim$50min &0.9175 & 30.20 &  0.1025 & - \\
\hline
\begin{tabular}[l]{@{}l@{}}InstantSplat~\cite{fan2024instantsplat}$^*$  \\(MASt3R MVS+3DGS)\end{tabular} 
 & -  & 14min56s & 0.5617 & 18.28 & 0.488 & 0.021  \\			
\hline
NeRF-mm \cite{wang2021nerf} & - &$\sim$13h33min& 0.5313 & 20.02 & 0.5450 & 0.035 \\
BARF \cite{lin2021barf} & GT K & $\sim$20h & 0.6075 & 23.42 &  0.5362 & 0.078 \\
NoPe-NeRF \cite{bian2023nope} & GT K & $\sim$30h & 0.7125 & 25.49& 0.4113 & 0.013\\
CF-3DGS \cite{fu2023colmap}& GT K & $\sim$2h20min & 0.9213 & 31.14 & 0.0859 & \textbf{0.004} \\
\hline
\textbf{Ours} & \textbf{-} & \textbf{26min20s} & \textbf{0.9347} & \textbf{31.59} & \textbf{0.0730} & \textbf{0.004} \\
\bottomrule
\end{tabular}}
\vspace{-2pt}
\caption{\small \textbf{Quantitative Evaluations on Tanks and Temples Dataset.} Our method achieves superior rendering quality and pose accuracy while requiring minimal training time and no camera parameters. (-) indicates no camera parameters required, GT K indicates known intrinsics, GT K \& Pose indicates both known intrinsics and extrinsics. \textbf{$^*$ InstantSplat cannot process dense views directly due to OOM (see Tab.~\ref{tab:oom}); thus, we adopt its chunk-by-chunk version, which yields inferior quality on long-sequence videos.}
}\label{tab:main_tt}
\end{table*}

\noindent\textbf{Metrics:}
We assess our approach on two key tasks: generating novel views and estimating camera poses. For the task of novel view synthesis, we evaluate performance using common metrics such as PSNR, SSIM~\cite{wang2004image}, and LPIPS~\cite{zhang2018unreasonable}. In terms of camera pose estimation, we evaluate Absolute Trajectory Error (ATE)~\cite{sturm2012benchmark} and utilize COLMAP-generated poses from all dense views as our ground-truth. While Relative Pose Error (RPE)~\cite{sturm2012benchmark} evaluates the local consistency of relative transformations between consecutive frames, it can be sensitive to discrepancies in intrinsic parameters such as focal length. ATE provides a more comprehensive measure of global trajectory accuracy and is better aligned with the goals of our method, which emphasizes globally consistent 3D reconstruction~\cite{sturm2012benchmark}. As such, we prioritize ATE as the primary metric for evaluating the poses of \name.

\noindent\textbf{Implementation Details:}
Our implementation is built on the PyTorch platform. During fragment registration, each fragment consists of $k=4$ frames. For depth map prediction, we utilize MASt3R with a resolution of 512 on the longer side. We run 200 iterations for key frame optimization. For hierarchical Gaussian alignment, we initialize each local Gaussian using the number of pixels within the fragment and train it for 200 steps. Camera poses are represented in quaternion format. For pair-wise merging,  the transformation matrix from key frame optimization is applied to the camera poses and Gaussian points of the subsequent local Gaussian. First, the camera poses are optimized with a learning rate of 1e-3 for 200 steps. Next, a mask is rendered to identify inadequately reconstructed regions, where new Gaussians are added. This process is repeated iteratively until a globally consistent Gaussian representation is achieved. We uniformly sample $\frac{1}{2}$ and $\frac{1}{4}$ training views on Tanks and Temples and CO3D-V2, respectively. All experiments were conducted on a single Nvidia A6000 GPU to maintain fair comparison.

\subsection{Quantitative Evaluations}
To quantitatively evaluate the quality of synthesized novel views, we present the results in Tab.~\ref{tab:main_tt} for the Tanks and Temples dataset and Tab.~\ref{tab:main_co3d} for the CO3D-V2 dataset. Baseline models were re-trained using their officially released code to ensure a fair comparison of training time. Compared to other self-calibrating radiance field methods, our approach achieves superior performance in terms of efficiency and rendering quality, which is largely thanks to our decoupled fragment registration and hierarchical alignment process. Compared to the most relevant baseline CF-3DGS~\cite{fu2023colmap}, we reduce $>$80\% training time yet get $>$0.012 LPIPS improvement on Tanks and Temples, and reduce $>$85\% training time yet get $>$0.12 LPIPS improvement on CO3D-V2 dataset. Note that our \name does not require any ground-truth camera parameters, making it more adaptable to scenes that do not have or fail to get precomputed intrinsics from COLMAP. Compared to NeRFmm~\cite{wang2021nerf}, which also does not need ground-truth camera parameters, our \name delivers much better quality and much less training time. Detailed per-scene breakdown results could be found in the Supplementary.

\begin{table*}
\centering
\resizebox{0.95\linewidth}{!}{
\begin{tabular}{lccccccc}
\toprule
& Camera Param. & Train Time & SSIM $\uparrow$ & PSNR $\uparrow$ & LPIPS $\downarrow$ & ATE $\downarrow$ \\\hline
COLMAP+3DGS & GT K \& Pose  & 15min44s  & 0.9211 & 32.26  & 0.1662  & - \\\hline
\begin{tabular}[l]{@{}l@{}}InstantSplat~\cite{fan2024instantsplat}$^*$  \\(MASt3R MVS+3DGS)\end{tabular} & -  & 19min3s & 0.6400 &  18.48 &  0.5355 & 0.045  \\
\hline
NeRF-mm \cite{wang2021nerf}& - & $\sim$17h22min & 0.4380 & 13.43 & 0.7058 & 0.061 \\
NoPe-NeRF \cite{bian2023nope}& GT K & $\sim$35h & 0.7030 & 25.54 & 0.5190 & 0.055 \\
CF-3DGS \cite{fu2023colmap} & GT K & $\sim$2h55min & 0.6821 & 22.98 & 0.3515 & 0.014 \\
\hline
\textbf{Ours} & - & \textbf{24min58s} & \textbf{0.8502} & \textbf{28.37} & \textbf{0.2237} & \textbf{0.012} \\
\bottomrule
\end{tabular}}
\vspace{-2pt}
\caption{\small \textbf{Quantitative Evaluations on CO3D-V2 Datasets.} (-) indicates no camera parameters required, GT K indicates known intrinsics. }\label{tab:main_co3d}
\vspace{-15pt}
\end{table*}


\begin{figure*}[t!]
  \centering
\includegraphics[width=0.88\linewidth]{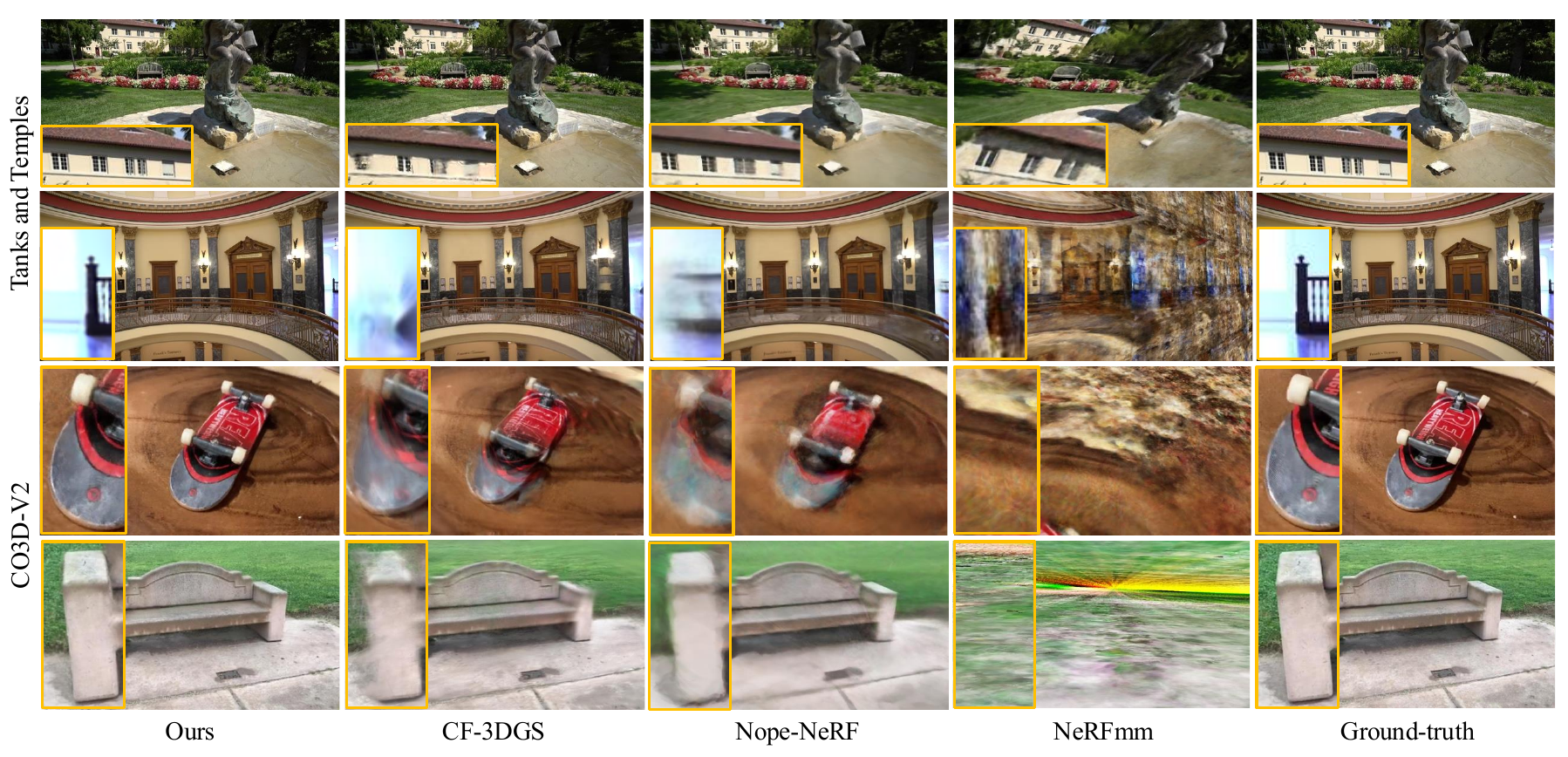}
  \vspace{-5pt}
\caption{\small \textbf{Visual Comparisons} between \name and other baselines. The insets highlight the details of renderings. \name achieves faithful 3D reconstruction, preserves better details, and alleviates incremental error in progressive learning.}\label{fig:tt_co3d}
\vspace{-15pt}
\end{figure*}

\subsection{Qualitative Evaluations}

As shown in Fig.~\ref{fig:tt_co3d}, for large-scale scenes from the Tanks and Temples dataset, thanks to the hierarchical design in \name, our method consistently produces sharper details among all test views, and preserves fine details that are well-optimized within each fragment. For the CO3D-V2 dataset, which includes 360-degree scenes with complex trajectories, achieving a globally consistent 3D reconstruction without any COLMAP initialization is even more challenging. Baselines that rely on monocular depth prediction to unproject images into point clouds often suffer from depth scale inconsistencies, making them fragile and prone to failure. Even CF-3DGS, which uses the more robust ZoeDepth monocular depth estimator~\cite{bhat2023zoedepth}, encounters severe failures on CO3D-V2. In contrast, \name leverages 3D geometry priors to achieve robust registration, making it highly adaptable and resilient in challenging settings. More results could be found in the Supplementary.


\subsection{Ablation Study}



Tab.~\ref{tab:ablation} reports the impact of various design choices on training time and reconstruction quality.

\vspace{-3pt}
\noindent\textbf{Local Fragment Registration.} Replacing it with direct MASt3R multi-view stereo initialization
\begin{wraptable}{r}{0.55\textwidth}
\vspace{-10pt}
  \centering
  \small
  \resizebox{0.98\linewidth}{!}{%
\begin{tabular}{lcccc}
\toprule
Model & Train Time& SSIM $\uparrow$ & PSNR $\uparrow$ & LPIPS $\downarrow$ \\ 
\midrule
\textbf{Ours}($k$ = 4, $\beta$ = 0.9) & 28min49s &0.8957  &30.02 &0.1745 \\ 
\midrule
Local: Use MASt3R MVS Init.~\cite{fan2024instantsplat} & 35min  & 0.8582 & 27.91  & 0.1768 \\
Global: Hierarchy $\rightarrow$ sequential & 53min12s  & 0.6969   & 20.22    & 0.3876\\ 
Global: Remove Key Frame Guidance & 35min42s  & 0.7629   & 24.35    & 0.3433 \\
\midrule
$k$ = 2 & 35min2s   & 0.8936 &29.77  & 0.2138   \\
$k$ = 4 & 28min49s & 0.8957  & 30.02 & 0.1745 \\
$k$ = 8  & 38min5s  & 0.8787 & 27.51 & 0.2743 \\
\midrule
$\beta$ = 0.5 & 23min18s  & 0.6529 & 18.50 & 	0.4457 \\
$\beta$ = 0.9  & 28min49s &0.8957  &30.02 &0.1745 \\
$\beta$ = 0.99  & 43min55s  & 0.8325   & 29.08    & 0.2691 \\
\bottomrule
\end{tabular}}
\vspace{-5pt}
\caption{\small\textbf{Ablation studies on 34\_1403\_4393 scene from CO3D-V2 Dataset.} $k$ denotes the number of frames in local fragment. $\beta$ denotes the rendering confidence threshold in Gaussian merging.}
\label{tab:ablation}
\vspace{-20pt}
\end{wraptable}
increases training time and lowers reconstruction quality, suggesting that outputs directly from MASt3R are less accurate and introduce errors to Gaussian optimization, especially in the challenging video setup.

\vspace{-3pt}
\noindent\textbf{Hierarchical Gaussian Alignment.} Removing our hierarchical alignment and instead adding local Gaussians sequentially (as in CF-3DGS~\cite{fu2023colmap}) prolongs training and hurts performance, showing the efficiency and accuracy gains from our hierarchical design.

\vspace{-3pt}
\noindent\textbf{Key Frame Guidance.} Omitting key frame guidance forces additional time for pose optimization without achieving optimal performance, showing the crucial role of key frames in stabilizing and accelerating the merging process.

\vspace{-3pt}
\noindent\textbf{Fragment Size \(k\).} A smaller \(k\) yields more precise intra-fragment registration but complicates fragment alignment, whereas a larger \(k\) reduces joint correspondences within fragment and degrades relative pose estimation, thus degrading the performance.

\vspace{-3pt}
\noindent\textbf{Confidence Threshold \(\beta\).} Setting \(\beta\) too low allows fewer Gaussians to merge, leading to under-reconstructed areas, while a high \(\beta\) merges too many Gaussians, slowing down training.

\vspace{-2pt}


\section{Conclusion and Limitations}

We presented VideoLifter, a framework for efficient 3D scene reconstruction from monocular videos without relying on pre-computed camera poses or known intrinsics. By leveraging learning-based stereo priors and a hierarchical alignment strategy with 3D Gaussian splatting, VideoLifter produces dense, globally consistent reconstructions with significantly reduced computational overhead compared to prior methods~\cite{fu2023colmap, bian2023nope}. A key limitation, shared with prior pose-free methods (e.g., CF-3DGS~\cite{fu2023colmap}), is the assumption of a pinhole camera model. Extending to more general camera models remains an important direction for future work.

\bibliographystyle{unsrt}
\bibliography{neurips_2025}


\newpage
\end{document}